\documentclass[10pt,twocolumn,letterpaper]{article}
\usepackage[pagenumbers]{cvpr}
\usepackage{microtype}
\newcommand{\method}{Paintress\xspace}
\newcommand{\bench}{Expedition-33\xspace}
\newcommand{\smallhead}[1]{\noindent\textbf{#1}\quad}
\setlist[itemize]{leftmargin=*,topsep=2pt,itemsep=1pt,parsep=0pt}
\setlist[enumerate]{leftmargin=*,topsep=2pt,itemsep=1pt,parsep=0pt}

\definecolor{cvprblue}{rgb}{0.21,0.49,0.74}
\usepackage[pagebackref,breaklinks,colorlinks,allcolors=cvprblue]{hyperref}
\title{Eidetic Code World Model}
\author{%
  Wenqing Tian\textsuperscript{1,2}\quad
  Zijian Zhang\textsuperscript{2}\quad
  Qiang Liu\textsuperscript{1,2}\quad
  Liang Wang\textsuperscript{1,2}\\[3pt]
  {\normalsize\textsuperscript{1}School of Artificial Intelligence, University of Chinese Academy of Sciences}\\
  {\normalsize\textsuperscript{2}Institute of Automation, Chinese Academy of Sciences}\\[2pt]
  {\small\href{https://gabrielinger.github.io/Eidetic-Code-World-Model/}{\raisebox{-0.32em}{\includegraphics[height=1.5em]{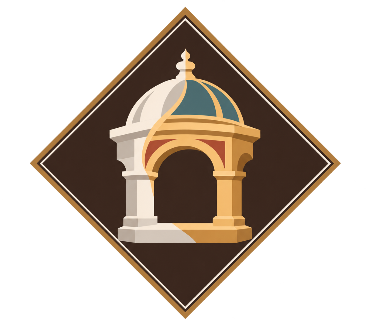}}\hspace{0.35em}Homepage}}%
}
\hypersetup{
  pdftitle={Eidetic Code World Model},
  pdfauthor={Wenqing Tian, Zijian Zhang, Qiang Liu, Liang Wang}
}
\begin{document}
\twocolumn[{
\renewcommand\twocolumn[1][]{#1}%
\maketitle
\centering
\includegraphics[width=\textwidth]{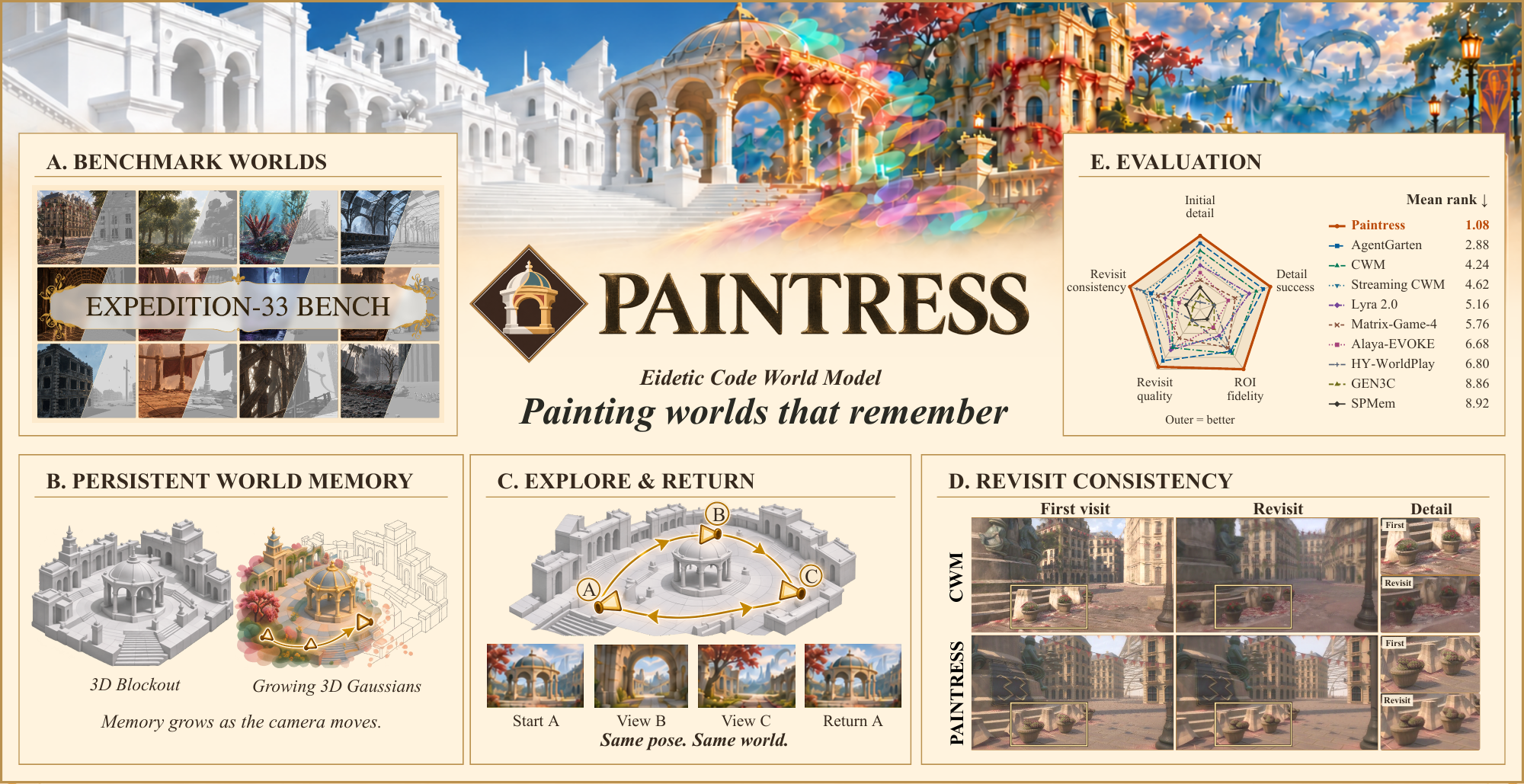}
\captionof{figure}{\textbf{\method: A world that remembers.} \textbf{(Top \& B)} \method couples an executable 3D blockout proxy with a persistent 3D Gaussian memory that grows during exploration. \textbf{(A \& C)} \bench evaluates whether models recover self-generated appearance at matched poses (Return A) after 15, 30, or 60\,s of target absence. \textbf{(D)} In Lumi\`ere (15\,s absence), CWM alters the monument banner and flowerpots upon revisit, whereas \method preserves its first-visit realization. \textbf{(E)} Across ten systems, \method ranks first overall across all five capability groups (Eq.~\ref{eq:meanrank}).}
\label{fig:teaser}
\par\medskip
}]
\begin{abstract}
Executable code can preserve a world's structural layout, yet video renderers frequently alter the appearance of previously visited places once they leave the camera view. We present \method, a code-conditioned world model that couples an executable geometric proxy with an independently parameterized 3D Gaussian appearance memory in a shared metric coordinate frame. While the proxy enforces structural and visibility constraints, 3D Gaussians capture the visual realization synthesized by emitted observations. To balance stable incremental fusion with consistent readout, \method separates writing from reading: newly emitted frames update a persistent raw Gaussian map under confidence-gated selection, whereas a cloned reader copy is globally refined against the observation archive and rendered with coverage masks to condition generation. We also introduce \bench, a benchmark for generated-appearance retention across 11 static scenes, five route families, and 15, 30, and 60\,s target absences (165 settings per system). Excluded from the opening image, targets are synthesized during a shared prefix and scored against each system's own first visit at matched return cameras. Across ten systems (1,650 videos), \method achieves 0.1220 ROI-LPIPS, 0.0613 MAE, and 52.12\% detail retention (vs.\ 0.2731, 0.1401, and 10.91\% for CWM), with lower discrepancy at 60 s (0.1630) than CWM, AgentGarten, and Lyra 2.0 achieve at 15 s.
\end{abstract}

\section{Introduction}
\label{sec:intro}
Returning from an expedition should reveal the world that was left behind. In generative video worlds, this requires remembering more than the coarse positions of walls and objects. Once a video renderer assigns a surface its specific color, material, and local pattern during exploration, those visual choices become part of the established world. If the underlying scene state remains static, a later revisit should recover that visual history. We study this requirement as \emph{generated-appearance retention}: preserving the visual realization synthesized during exploration, particularly for regions absent from the initial conditioning image.

Code-conditioned world models provide a principled foundation for structural persistence by separating world logic from visual synthesis~\cite{cwm,streamingcwm,agentgarten,matrixgame4,coder,mallea}. To keep programmatic state tractable to simulate, executable worlds typically render lightweight, low-poly whitebox proxies~\cite{cwm,agentgarten,matrixgame4,mallea} while delegating rich visual detail to a video diffusion renderer. A coarse proxy, however, is one-to-many: it specifies \emph{where} a surface lies, but not \emph{which} appearance the renderer previously synthesized. Conditioning on an opening anchor image and a sliding window of recent frames~\cite{cwm,streamingcwm} preserves views near the start or immediate past, but does not by itself guarantee preservation once newly explored regions leave the context horizon. In our observations (Fig.~\ref{fig:teaser}D), even when the proxy keeps the architectural layout intact, the renderer alters the monument banner, facade texture, and flowerpots upon revisit.

Recent works augment world models with long-term memory by registering generated pixels onto whitebox surfaces~\cite{coder,mallea}, reprojecting depth-derived point caches~\cite{gen3c,spmem,evoke}, retrieving historical observations via spatial or geometric cues~\cite{hyworldplay,lyra2}, or updating 3D Gaussian maps in place~\cite{l3gworld}. Coupling persistent appearance with an executable proxy raises two design questions. First, \emph{how should appearance memory be parameterized when synthesized details depart from coarse proxy geometry?} Because whitebox proxies approximate ornate facades or foliage with simplified primitives, we hypothesize that decoupling 3D appearance primitives from strict proxy-surface binding better accommodates synthesized geometry beyond the whitebox, provided reconstruction remains metrically anchored to the proxy. Second, \emph{how should memory reconcile selective incremental accumulation with globally consistent readout?} Updating a single persistent map in place at every step risks propagating global smoothing changes into the confidence criteria that govern subsequent multi-view fusion.

\method addresses both questions through a \emph{Dual-World Architecture} coupling an executable 3D blockout proxy with an independently parameterized RGB 3D Gaussian memory in a shared metric coordinate frame (Fig.~\ref{fig:teaser}B). Emitted RGB frames are reconstructed via VGGT-$\Omega$~\cite{vggtomega} and registered to known control cameras with soft proxy-depth regularization. \method also separates memory writing from reading: a persistent raw Gaussian map receives causal, confidence-gated incremental fusion, whereas a separate reader copy is cloned and globally refined against the observation archive. The proxy and the reader copy are then rendered at the same target cameras to supply structural and coverage-gated appearance conditions.

To make generated-appearance retention measurable, we introduce \bench (Fig.~\ref{fig:teaser}A, C). Complementing benchmarks that evaluate within-rollout mined pairs~\cite{r2m}, fixed references~\cite{worldmembench}, timestamped program events~\cite{prowbench}, or short-horizon draft realization~\cite{worlddraft}, \bench isolates \emph{first-visit-conditioned retention} of targets excluded from the opening image across 165 settings per system (11 scenes $\times$ 5 routes $\times$ 15/30/60\,s absence). Each system synthesizes the target during a shared prefix; the protocol restores this state for every absence branch, and each return is scored against the system's \emph{own} first visit at matched return cameras across appearance, detail retention, visibility timing, and visual quality. Across ten systems (1,650 videos), \method ranks first overall (Fig.~\ref{fig:teaser}E), reducing ROI-LPIPS by 55.3\% and MAE by 56.2\% relative to CWM while maintaining lower discrepancy at 60\,s (0.1630) than CWM, AgentGarten, and Lyra 2.0 achieve at 15\,s ($\ge 0.2579$).

Our contributions are threefold:
\begin{itemize}
\item \textbf{Paintress:} A dual-world architecture coupling executable proxy geometry with RGB 3D Gaussian memory via confidence-gated raw fusion, a separately refined reader copy, and coverage-aware conditioning.
\item \textbf{Expedition-33:} A 165-setting diagnostic benchmark for self-generated appearance retention with target-absent initialization, state-restored absence branches, and decoupled appearance, detail, timing, and quality metrics.
\item \textbf{Empirical characterization:} A 1,650-video comparison across ten systems, duration-specific degradation analyses, and stage-wise runtime breakdowns.
\end{itemize}

\section{Related Work}
\label{sec:related}

\begin{table*}[t]
\centering
\caption{\textbf{Comparison of structural control and appearance-memory interfaces.} ``Exec.\ proxy'' denotes a code- or engine-executed geometric world supplying structural conditions. $^\dagger$ discussed in related work but not included in quantitative evaluation.}
\begingroup
\footnotesize
\renewcommand{\arraystretch}{1.12}
\setlength{\tabcolsep}{0pt}
\providecommand{\cmark}{\textcolor[HTML]{214B3F}{\textbf{$\checkmark$}}}
\providecommand{\namark}{\textcolor[HTML]{858A7B}{---}}
\begin{tabular*}{\textwidth}{@{\extracolsep{\fill}}lcll@{}}
\toprule
\textbf{Method} & \textbf{Exec.\ proxy} & \textbf{Appearance history / memory} & \textbf{Appearance reuse} \\
\midrule
CWM~\cite{cwm} & \cmark & Opening image + recent RGB frames & Anchor and overlapping-frame conditioning \\
Streaming CWM~\cite{streamingcwm} & \cmark & Opening image + sliding KV cache & Causal attention over bounded history \\
AgentGarten~\cite{agentgarten} & \cmark & Reference/sink prefix + recent KV cache & Attention to reference and recent history \\
Matrix-Game 4.0~\cite{matrixgame4} & \cmark & Spatial-memory frames + reference/history cache & Spatial references and cached attention \\
CoDeR$^\dagger$~\cite{coder} & \cmark & Surface-registered appearance & Re-rendering registered whitebox surfaces \\
Mallea$^\dagger$~\cite{mallea} & \cmark & Object-local surface textures & State- and view-compatible masked projection \\
\midrule
HY-WorldPlay~\cite{hyworldplay} & \namark & Historical frame bank & Spatiotemporal retrieval and temporal reframing \\
Lyra 2.0~\cite{lyra2} & \namark & Historical frames + per-frame 3D cache & Geometry-aware retrieval and correspondence routing \\
GEN3C~\cite{gen3c} & \namark & RGB point-cloud caches & Target-view RGB warps and coverage masks \\
SPMem~\cite{spmem} & \namark & Static point map + historical frames & Point-map rendering and historical cross-attention \\
Alaya-EVOKE~\cite{evoke} & \namark & Camera-indexed geometry bank & Co-visible source projection and visibility masking \\
L3G-World$^\dagger$~\cite{l3gworld} & \namark & RGB/latent 3D Gaussian memory & Direct latent rendering and opacity conditioning \\
\midrule
\textbf{Paintress (Ours)} & \cmark & \textbf{Raw RGB 3DGS + observation archive} & \textbf{Refined-copy RGB rendering and coverage-aware conditioning} \\
\bottomrule
\end{tabular*}
\endgroup

\label{tab:related_methods}
\end{table*}

\begin{table}[t]
\centering
\caption{\textbf{Comparison of world-memory benchmarks.} \textbf{Exec.\ Proxy}: executable structure control; \textbf{In-Rollout Ref.}: returns compared against earlier generated views; \textbf{Target Absent}: target excluded from $I_0$; \textbf{Restored Tiers}: 15/30/60\,s branches from a shared prefix state.}
\label{tab:related_benchmarks}
\begingroup
\footnotesize
\renewcommand{\arraystretch}{1.12}
\setlength{\tabcolsep}{0pt}
\providecommand{\cmark}{\textcolor[HTML]{214B3F}{\textbf{$\checkmark$}}}
\providecommand{\namark}{\textcolor[HTML]{858A7B}{---}}
\begin{tabular*}{\columnwidth}{@{\extracolsep{\fill}}lccccr@{}}
\toprule
& \textbf{Exec.} & \textbf{In-Rollout} & \textbf{Target} & \textbf{Restored} & \\[-1.5pt]
\textbf{Benchmark} & \textbf{Proxy} & \textbf{Ref.} & \textbf{Absent} & \textbf{Tiers} & \textbf{Scale} \\
\midrule
R2M-Bench~\cite{r2m} & \namark & \cmark & \namark & \namark & 300 \\
WorldMemBench~\cite{worldmembench} & \namark & \namark & \namark & \namark & 260 \\
PROWBench~\cite{prowbench} & \cmark & \namark & \namark & \namark & 170 \\
WorldDraft~\cite{worlddraft} & \cmark & \namark & \namark & \namark & 432 \\
\midrule
\textbf{Expedition-33} & \cmark & \cmark & \cmark & \cmark & \textbf{165} \\
\bottomrule
\end{tabular*}
\endgroup

\end{table}

\paragraph{Executable worlds and stored appearance.}
Code- and engine-driven world models decouple state simulation from visual synthesis by rendering low-poly geometric proxies to condition video generation~\cite{cwm,streamingcwm,agentgarten,matrixgame4,coder,mallea}. CWM~\cite{cwm} pairs an executable proxy with an opening anchor and a 34-frame overlapping window, while Streaming CWM~\cite{streamingcwm} uses causal chunks with a sliding KV cache. AgentGarten~\cite{agentgarten} refines code-world construction via feedback-driven repair and evolves agents through inherited playbooks, conditioning its renderer on proxy depth or camera-space normals alongside a reference/sink prefix and recent KV cache. Matrix-Game~4.0~\cite{matrixgame4} combines engine-maintained low-poly proxies with spatial-memory reference frames and cached history. To persist appearance on proxy geometry, CoDeR~\cite{coder} registers generated pixels onto whitebox surfaces for re-rendering, and Mallea~\cite{mallea} stores object-local surface textures with state-, view-, and visibility-compatible projection. Rather than binding appearance primitives to coarse whitebox surfaces, \method reconstructs emitted RGB observations as 3D Gaussians in the proxy's metric coordinate frame—regularized by proxy depth—and maintains them via a raw fusion parent and a separately refined reader copy (Table~\ref{tab:related_methods}).

\paragraph{Spatial memories for generative video worlds.}
Camera- and action-conditioned world models preserve long-horizon consistency via historical frame banks or explicit 3D representations~\cite{hyworldplay,lyra2,gen3c,spmem,evoke,l3gworld}. WorldPlay~\cite{hyworldplay} retrieves historical frames via spatiotemporal cues and temporal reframing, while Lyra~2.0~\cite{lyra2} pairs historical frames with a per-frame 3D cache for geometry-aware retrieval and correspondence routing. Among explicit 3D caches, GEN3C~\cite{gen3c} warps RGB point-cloud caches with coverage masks, SPMem~\cite{spmem} combines a static point map with historical cross-attention, and Alaya-EVOKE~\cite{evoke} maintains a camera-indexed geometry bank with co-visible projection and visibility masking. Contemporaneous work L3G-World~\cite{l3gworld} stores RGB and VAE-latent attributes in a persistent 3D Gaussian memory and directly renders latent features for video conditioning. Our focus is coupling an executable geometric proxy with RGB Gaussian appearance memory, while preserving a raw fusion map and deriving a separately refined reader for generation.

\paragraph{Evaluating appearance persistence.}
Recent benchmarks examine distinct facets of memory and control in video world models (Table~\ref{tab:related_benchmarks}). R2M-Bench~\cite{r2m} evaluates revisit-selective consistency across 300 trajectories by pairing earlier visits with later returns within the same rollout alongside non-revisit and short-range controls. WorldMemBench~\cite{worldmembench} tests static retention and off-screen evolution across 260 cases with repeated returns to fixed references. For executable worlds, PROWBench~\cite{prowbench} evaluates adherence to programmatic entity states and timestamped events across 170 cases, while WorldDraft~\cite{worlddraft} crosses draft granularity with appearance alignment across 432 clips and gates attribute persistence by anchor correctness. \bench complements these suites by combining within-rollout generated references with executable proxy control across 165 settings per system: excluded from $I_0$, the target is synthesized during a shared prefix and evaluated after state-restored 15, 30, and 60\,s absences at matched return poses.

\section{The Expedition-33 Benchmark}
\label{sec:benchmark}
\bench evaluates \emph{generated-appearance retention}: whether a world model preserves the visual realization it synthesizes during exploration while the executable scene state remains static. As shown in Fig.~\ref{fig:protocol}, the benchmark comprises 11 evaluation scenes spanning streets, woodland, stations, interiors, and vertically layered structures (plus one 15-sample development scene, Stone Wave Cliffs). Crossing the 11 scenes with three target-absence intervals (15, 30, and 60\,s) defines 33 scene--interval settings, and five route families per setting (Routes A--E, logged as R1--R5) yield $11\times3\times5=165$ evaluation samples per system.

\subsection{Asset Construction and Revisit Protocol}
\label{sec:protocol}
\begin{figure*}[t]
\centering
\includegraphics[width=\textwidth]{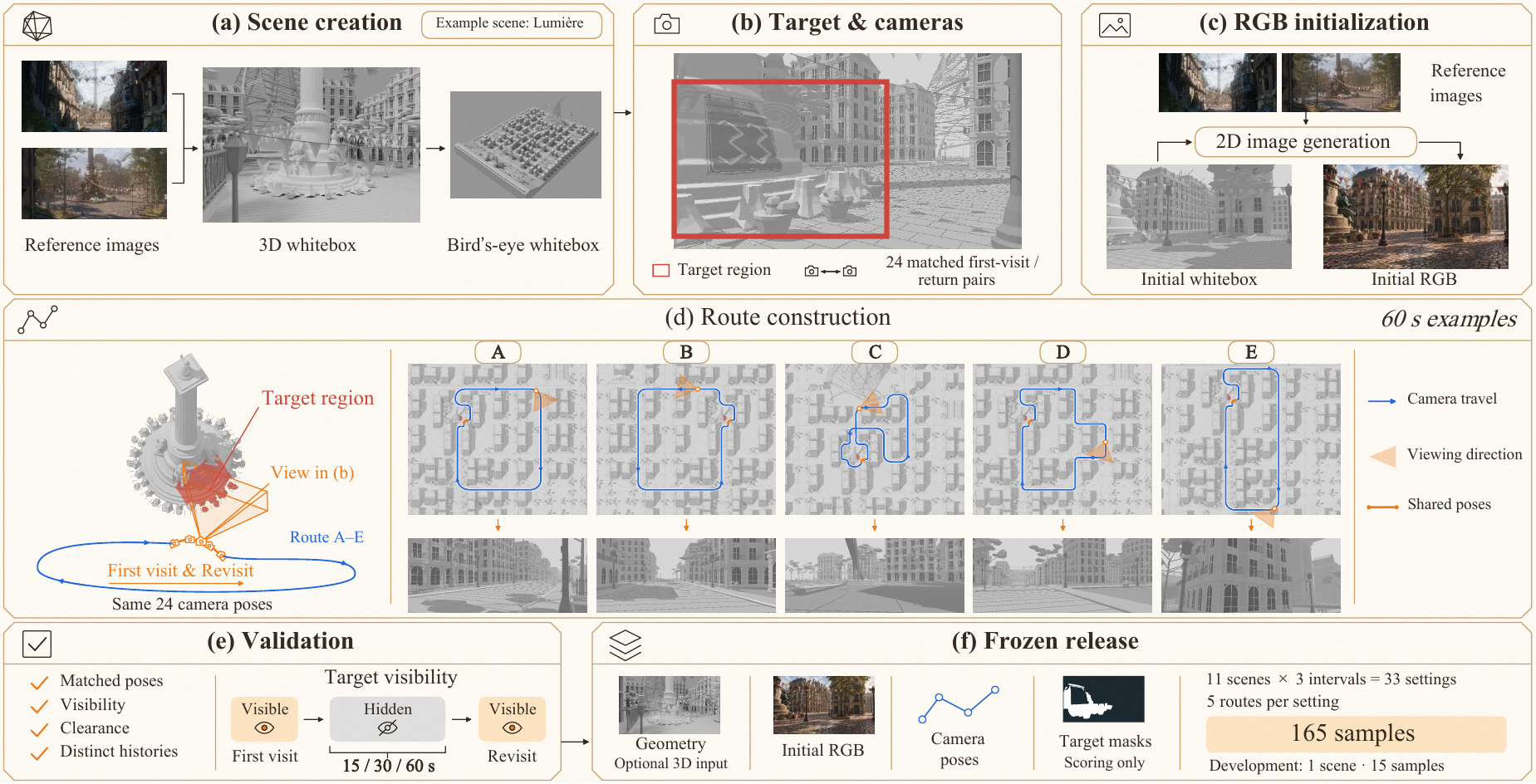}
\caption{\textbf{Construction and protocol of \bench.} \textbf{(a--c)} Authored 3D whiteboxes define target regions, 24 matched first-visit/return poses, and target-absent opening images $I_0$. \textbf{(d)} Five route families (A--E) share the 24 reference poses while traversing distinct paths during 15, 30, or 60\,s target absences. \textbf{(e--f)} Validated trajectories are released across 165 evaluation samples with scoring-only target masks.}
\label{fig:protocol}
\end{figure*}

Each scene is authored as an executable 3D whitebox guided by reference images, defining a target region and 24 matched first-visit/return camera poses (Fig.~\ref{fig:protocol}a--b). To prevent models from copying the opening frame, the initial whitebox view is selected so that the evaluation target lies strictly outside the frustum, and 2D image generation produces the opening RGB image $I_0$ (Fig.~\ref{fig:protocol}c). During evaluation, a shared prefix trajectory guides the camera from $I_0$ to discover and synthesize the target across the 24 reference poses. At the end of this prefix, the protocol checkpoints and restores the model's complete continuation state—including recent frames, random state, spatial memory, and retrieval caches—before branching into five distinct exploration routes (A--E; Fig.~\ref{fig:protocol}d) that keep the target out of view for a prescribed absence interval $g=t^*_{\mathrm{reappear}}-t^*_{\mathrm{last\ visible}}\in\{15,30,60\}\,\mathrm{s}$. All trajectories are validated for pose matching, visibility transitions, obstacle clearance, and distinct histories, and released with proxy geometry, $I_0$, camera poses, and scoring-only target masks over the 24 matched return poses (1.0\,s at 24\,FPS; Fig.~\ref{fig:protocol}e--f).

\subsection{Evaluation Metrics}
\label{sec:metrics}
To prevent camera failure or oversmoothed rendering from confounding evaluation~\cite{r2m}, \bench measures four complementary dimensions:

\smallhead{Target appearance and whole-frame consistency.}
Each return frame is compared against the method's own generated first visit at the paired pose on $640\times360$ content images (RGB in $[0,1]$). We compute MAE, MSE, PSNR, and local SSIM~\cite{ssim} within the fixed target mask, and ROI-LPIPS~\cite{lpips} on the fixed target bounding box and surrounding context. Following R2M-Bench~\cite{r2m}, we also record whole-frame RGB MAE on revisit ($R$), non-revisit ($B$), and short-range ($S$) pairs, with per-record memory gain $\mathrm{MG}=B-R$.

\smallhead{Detail establishment and retention.}
Each target defines $K_i$ attribute slots (\eg, color, material, pattern) whose reference values are established during the generated first visit. With $a_{ij}\in\{0,1\}$ indicating first-visit recognizability of slot $j$ and $b_{ij}\in\{0,1\}$ indicating return preservation, we define:
\begin{equation}
\textstyle C_i=\frac{1}{K_i}\sum_j a_{ij},\quad
D_i=\frac{1}{K_i}\sum_j a_{ij}b_{ij},\quad
R_{\mathrm{cond},i}=\frac{D_i}{C_i},
\label{eq:detail}
\end{equation}
measuring initial establishment ($C_{\mathrm{first}}$ across unique prefixes), overall retention ($D$), and conditional retention ($R_{\mathrm{cond}}$, defined when $\sum_j a_{ij}>0$).

\smallhead{Visibility timing and visual quality.}
Sampled visibility checks bracket observed departure and first reappearance timestamps, yielding measurement bounds $[\text{lower}, \text{upper}]$ for boundary errors, absence-duration errors, and signed bias $\Delta T$, alongside the percentage of settings with a bounded interval (\emph{Estimable \%}). A VLM judge (Gemini 4 Argon) separately assesses target existence, attribute readability, sharpness, artifact absence, and structural plausibility at first visit and return. The same judge labels $a_{ij}$ and $b_{ij}$ and performs the sampled visibility checks.

\subsection{Support, Aggregation, and Overall Ranking}
\label{sec:aggregation}
All systems are evaluated on all 165 samples ($55$ per interval). For model $m$, metric $s$, and interval $g\in\{15,30,60\}$ with valid set $\mathcal{V}^s_{m,g}$, the macro-average equally weights the three interval means:
\begin{equation}
\textstyle \overline{s}_m=\frac{1}{3}\sum_{g\in\{15,30,60\}}
\frac{1}{|\mathcal{V}^s_{m,g}|}\sum_{i\in\mathcal{V}^s_{m,g}}s_{m,i}.
\label{eq:aggregation}
\end{equation}
Unconditional metrics include all $|\mathcal{V}^s_{m,g}|=55$ records per interval; conditional metrics average over eligible records and are denoted `---' if any interval has $|\mathcal{V}^s_{m,g}|=0$.

To summarize overall performance across the $M=10$ systems (Fig.~\ref{fig:teaser}E), we rank systems from $1$ (best) to $10$ (worst) on each aggregate metric (averaging tied ranks) across five equally weighted groups $\{G_g\}_{g=1}^5$: initial detail ($C_{\mathrm{first}}$, $|G_1|=1$), detail success ($D$, $|G_2|=1$), ROI fidelity (ROI-LPIPS, MAE, MSE, PSNR, SSIM; $|G_3|=5$), revisit quality (five return quality dimensions; $|G_4|=5$), and revisit consistency ($R$, $|G_5|=1$):
\begin{equation}
\textstyle \operatorname{MeanRank}(m)
=\frac{1}{5}\sum_{g=1}^{5}
\left(\frac{1}{|G_g|}\sum_{j\in G_g}r_{m,j}\right),
\label{eq:meanrank}
\end{equation}
with radar radii in Fig.~\ref{fig:teaser}E plotted as $11 - \frac{1}{|G_g|}\sum_{j\in G_g}r_{m,j}$.

\section{Paintress: A Dual-World Architecture}
\label{sec:method}

\begin{figure*}[t]
\centering
\includegraphics[width=\textwidth]{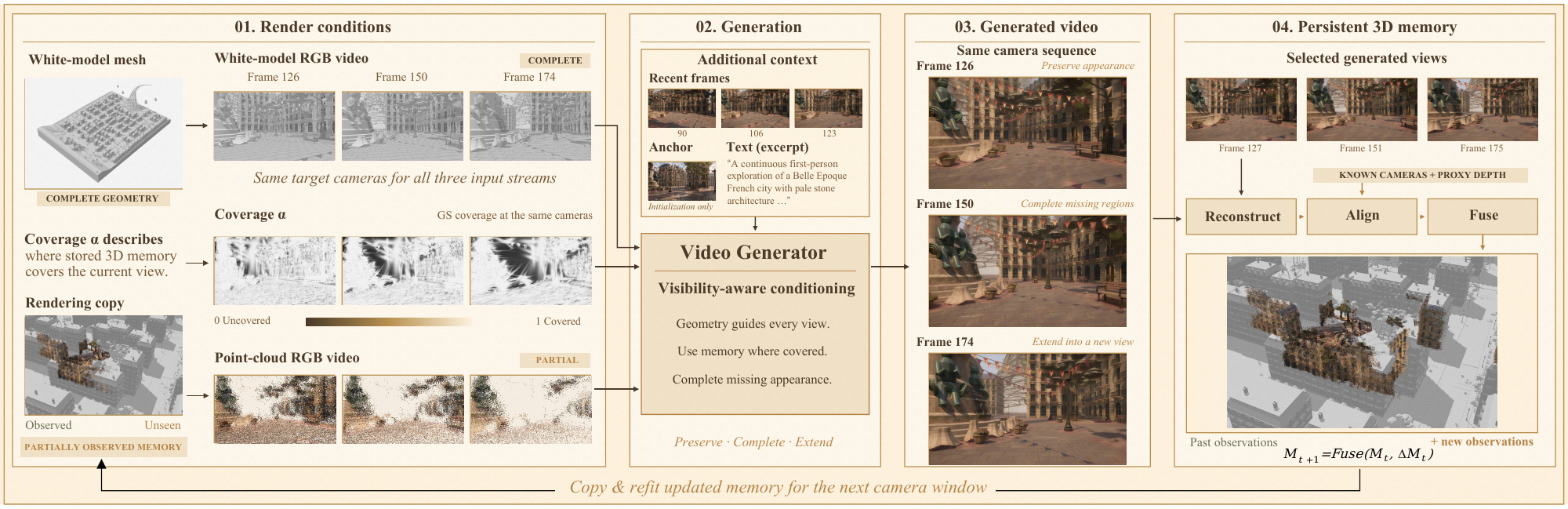}
\caption{\textbf{Overview of the Paintress Dual-World Architecture.} An executable geometric world and a persistent 3D Gaussian memory share a metric coordinate frame and are rendered at target cameras $C_k$. A cloned reader copy $M^r_{k-1}$ is refined against the observation archive $\mathcal{D}_{<k}$ and rendered into RGB appearance $A_k$ and coverage $\alpha_k$ to condition H3 alongside the gray proxy $P_k$, while newly emitted frames update the persistent raw map $M^w_k$ via VGGT-$\Omega$, $\mathrm{Sim}(3)$ alignment, local fitting, and confidence-gated fusion.}
\label{fig:method}
\end{figure*}

\subsection{Dual-World Formulation}
\method maintains two synchronized representations in a shared metric coordinate frame (Fig.~\ref{fig:method}): an executable code state $S$ providing structural layout and visibility, and an independently parameterized 3D Gaussian memory storing the visual appearance synthesized during exploration. Because whitebox proxies simplify complex structures (\eg, ornate facades, banners, or foliage) into coarse primitives, strictly binding stored appearance to proxy surfaces risks geometric mismatch when synthesized details depart from the whitebox plane, whereas unanchored reconstruction suffers from scale and pose drift. \method addresses both issues by allowing 3D Gaussians to occupy positions not bound to proxy surfaces while anchoring their registration and depth regularization to the executable proxy.

For continuation window $k\ge 1$ with target cameras $C_k$, the executable world renders a gray proxy video $P_k=\mathcal{R}_{\mathrm{code}}(S,C_k)$. From the persistent raw Gaussian map $M^w_{k-1}$ and the archive $\mathcal{D}_{<k}$ of selected past RGB frames, cameras, and proxy depths, we derive a refined reader map $M^r_{k-1}$ and render target-view appearance $A_k$ and opacity $\alpha_k$:
\begin{equation}
M^r_{k-1} = F_k(M^w_{k-1},\mathcal{D}_{<k}), \quad
(A_k,\alpha_k) = \mathcal{R}_{\mathrm{GS}}(M^r_{k-1},C_k).
\label{eq:method:read}
\end{equation}
Conditioned on $P_k$, $(A_k,\alpha_k)$, the 34-frame recent context $H_k$, and prompt $T$, the generator $G_\theta$ synthesizes the next window:
\begin{equation}
\widehat{V}_k=G_\theta(P_k,A_k,\alpha_k,H_k,T).
\label{eq:method:generate}
\end{equation}
The opening image $I_0$ conditions only cold start ($k=0$).

\subsection{Causal Gaussian Memory Construction}
After emitting window $k$, we select 16 non-overlapping frames $\mathcal{Q}^{\mathrm{RGB}}_k$ to update memory. VGGT-$\Omega$~\cite{vggtomega} predicts relative poses, dense depth, and pixelwise confidence from $\mathcal{Q}^{\mathrm{RGB}}_k$. A single $\mathrm{Sim}(3)$ transformation aligns predicted camera centers and orientations to known control cameras $C_k^q$, using robust residuals against proxy depth $D_k^q$ to constrain metric scale (with a fallback to known extrinsics and robust depth scaling if positive alignment fails). $D_k^q$ regularizes scale without overwriting predicted depth, preserving fine synthesized geometry.

The aligned points initialize anisotropic 3D Gaussians~\cite{gaussiansplatting} with degree-zero spherical-harmonic colors, optimized under fixed cameras $C_k^q$ for 1,500 steps via:
\begin{equation}
\mathcal{L}_{\mathrm{fit}}=0.8\mathcal{L}_{1}
+0.2(1-\operatorname{SSIM})
+0.2\mathcal{L}_{\mathrm{depth}}
+0.01\overline{o},
\label{eq:method:fit}
\end{equation}
where $\mathcal{L}_{\mathrm{depth}}$ penalizes relative depth error against $D_k^q$ and $\overline{o}$ is mean opacity. The fitted candidate $\widetilde{M}_k=\operatorname{Fit}(\operatorname{Align}(\operatorname{Omega}(\mathcal{Q}^{\mathrm{RGB}}_k);C_k^q,D_k^q),\mathcal{Q}_k)$ is merged into $M^w_{k-1}$ via a $0.05\,\mathrm{m}$ spatial selection grid $U(\cdot,\cdot)$ that retains reliable cells, appends uncovered cells, and replaces weak cells only under stronger multi-view confidence:
\begin{equation}
M^w_k = U(M^w_{k-1},\widetilde{M}_k), \qquad
\mathcal{D}_{<k+1} = \mathcal{D}_{<k}\cup\mathcal{Q}_k.
\label{eq:method:write}
\end{equation}

\subsection{Read/Write Separation and Conditioning}
While confidence-gated fusion $U$ preserves reliable local geometry, independently fitted chunks can exhibit subtle cross-window seams. Globally optimizing a single persistent map in place at every step, however, would shift Gaussian positions and opacities, invalidating the spatial support metadata governing subsequent writes. \method therefore decouples reading from writing: $F_k$ clones $M^w_{k-1}$ and refines the copy for 1,500 steps across $\mathcal{D}_{<k}$ to produce $M^r_{k-1}$ for rendering, discarding the clone afterward so $M^w_{k-1}$ remains the uncorrupted fusion parent.

We instantiate $G_\theta$ on MiniMax H3~\cite{h3}, packing encoded $P_k$ and $A_k$ reference blocks with target latents. To distinguish unvisited black space from dark surfaces, opacity $\alpha_k$ (1) gates GS latents, (2) masks invalid GS keys ($\alpha_k < 0.5$) in the Qwen branch, and (3) drops uncovered GS keys from H3 attention. Each 124-frame native window (24\,FPS) conditions on 34 prefix frames (10 latent slices) to emit 90 new frames (3.75\,s) in four flow-matching updates without CFG, using rank-16 attention LoRA~\cite{lora} ($\alpha=16$) with base weights frozen.

\begin{table*}[t]
\centering
\caption{\textbf{RQ1: Appearance consistency and detail retention across all 165 settings.} Macro-averages across the 15, 30, and 60\,s absence tiers, with every system scored against its \emph{own} generated first visit. `---' denotes an undefined aggregate under the stated support and aggregation rules (Sec.~\ref{sec:aggregation}). \textbf{Bold} marks the best result and \underline{underline} the second best.}
\label{tab:e33_main_v06}
\footnotesize
\renewcommand{\arraystretch}{1.08}
\setlength{\tabcolsep}{0pt}
\begin{tabular*}{\textwidth}{@{\extracolsep{\fill}}lrrrrrrrrrr@{}}
\toprule
& \multicolumn{3}{c}{\textbf{Detail Retention}} & \multicolumn{5}{c}{\textbf{Target-Region Appearance}} & \multicolumn{2}{c}{\textbf{Whole-Frame}} \\
\cmidrule(lr){2-4} \cmidrule(lr){5-9} \cmidrule(l){10-11}
System & $C_{\mathrm{first}}$ (\%)$\uparrow$ & $D$ (\%)$\uparrow$ & $R_{\mathrm{cond}}$ (\%)$\uparrow$ & ROI-LPIPS$\downarrow$ & MAE$\downarrow$ & MSE$\downarrow$ & PSNR (dB)$\uparrow$ & SSIM$\uparrow$ & $R\downarrow$ & MG$\uparrow$ \\
\midrule
\textbf{Paintress (Ours)} & \textbf{86.06} & \textbf{52.12} & \textbf{57.58} & \textbf{0.1220} & \textbf{0.0613} & \textbf{0.0060} & \textbf{24.35} & \textbf{0.5511} & \textbf{0.0739} & \textbf{0.1068} \\
CWM~\cite{cwm} & 60.00 & 10.91 & 18.79 & 0.2731 & \underline{0.1401} & \underline{0.0342} & 15.75 & 0.3415 & 0.1644 & 0.0305 \\
Streaming CWM~\cite{streamingcwm} & 50.30 & 3.03 & 6.06 & 0.2777 & 0.1583 & 0.0442 & 14.71 & 0.2289 & 0.1465 & 0.0116 \\
AgentGarten~\cite{agentgarten} & \underline{67.88} & \underline{11.52} & 16.97 & 0.2699 & 0.1420 & 0.0492 & 15.96 & 0.3971 & 0.1536 & 0.0302 \\
GEN3C~\cite{gen3c} & 10.30 & 0.00 & 0.00 & 0.3543 & 0.1954 & 0.0649 & 12.97 & 0.4075 & 0.1932 & $-0.0014$ \\
SPMem~\cite{spmem} & 10.91 & 0.00 & --- & 0.4374 & 0.2439 & 0.0970 & 11.22 & 0.3140 & 0.1931 & --- \\
Alaya-EVOKE~\cite{evoke} & 39.39 & 0.24 & 0.67 & 0.3266 & 0.1846 & 0.0639 & 13.10 & 0.3106 & 0.1749 & 0.0120 \\
HY-WorldPlay~\cite{hyworldplay} & 0.00 & 0.00 & --- & \underline{0.1810} & 0.1474 & 0.0462 & \underline{18.55} & 0.3195 & \underline{0.1272} & \underline{0.0342} \\
Lyra 2.0~\cite{lyra2} & 43.64 & 7.88 & \underline{26.06} & 0.2673 & 0.1710 & 0.0548 & 13.62 & 0.2433 & 0.1596 & \underline{0.0342} \\
Matrix-Game-4~\cite{matrixgame4} & 18.79 & 0.61 & 3.64 & 0.3244 & 0.1500 & 0.0457 & 15.12 & \underline{0.4541} & 0.1560 & 0.0234 \\
\bottomrule
\end{tabular*}
\end{table*}

\begin{table*}[t]
\centering
\caption{\textbf{RQ3: Target-visibility timing adherence and estimability.} Macro-averages across the 15, 30, and 60\,s tiers. \emph{Estimable} is the percentage of videos with a bounded absence interval; brackets denote sampled measurement bounds in seconds (or \%), and `---' denotes an undefined aggregate under the stated support and aggregation rules.}
\label{tab:e33_timing_v05}
\footnotesize
\renewcommand{\arraystretch}{1.08}
\setlength{\tabcolsep}{0pt}
\begin{tabular*}{\textwidth}{@{\extracolsep{\fill}}lrrrrrrr@{}}
\toprule
System & Estimable (\%)$\uparrow$ & Departure Err.$\downarrow$ & Reappear.\ Err.$\downarrow$ & Boundary Err.$\downarrow$ & Duration Err.$\downarrow$ & Rel.\ Dur.\ Err.$\downarrow$ & Signed Bias $\Delta T$ \\
\midrule
\textbf{Paintress (Ours)} & \textbf{100.00} & \textbf{[0.25, 0.97]} & \textbf{[1.83, 5.22]} & \textbf{[1.04, 3.09]} & \textbf{[1.69, 5.75]} & \textbf{[5.45\%, 18.18\%]} & \textbf{[$-$1.93, $+$2.34]} \\
CWM~\cite{cwm} & 44.85 & [0.62, 1.25] & [8.42, 11.45] & [4.56, 6.33] & [7.85, 11.39] & [17.58\%, 29.70\%] & [$-$6.17, $-$2.44] \\
Streaming CWM~\cite{streamingcwm} & 32.73 & [1.05, 1.95] & [7.27, 9.73] & [4.06, 5.60] & [6.87, 9.90] & [20.61\%, 32.12\%] & [$-$1.65, $+$1.94] \\
AgentGarten~\cite{agentgarten} & \underline{68.48} & \underline{[0.36, 0.83]} & [3.39, 7.04] & \underline{[1.87, 3.92]} & \underline{[3.18, 7.11]} & \underline{[10.30\%, 24.24\%]} & [$-$0.07, $+$4.41] \\
GEN3C~\cite{gen3c} & 0.00 & [2.63, 3.47] & --- & --- & --- & --- & --- \\
SPMem~\cite{spmem} & 0.00 & --- & --- & --- & --- & --- & --- \\
Alaya-EVOKE~\cite{evoke} & 1.21 & [0.96, 1.89] & --- & --- & --- & --- & --- \\
HY-WorldPlay~\cite{hyworldplay} & 0.00 & --- & --- & --- & --- & --- & --- \\
Lyra 2.0~\cite{lyra2} & 26.06 & [3.55, 4.73] & \underline{[3.28, 28.82]} & [2.86, 16.17] & [4.03, 30.73] & [12.12\%, 87.88\%] & [$-$30.73, $-$0.64] \\
Matrix-Game-4~\cite{matrixgame4} & 5.45 & [1.81, 2.54] & [5.75, 6.80] & [2.93, 3.64] & [5.56, 7.00] & [19.39\%, 25.45\%] & [$+$5.56, $+$7.00] \\
\bottomrule
\end{tabular*}
\end{table*}

\begin{figure*}[t]
\centering
\includegraphics[width=\textwidth]{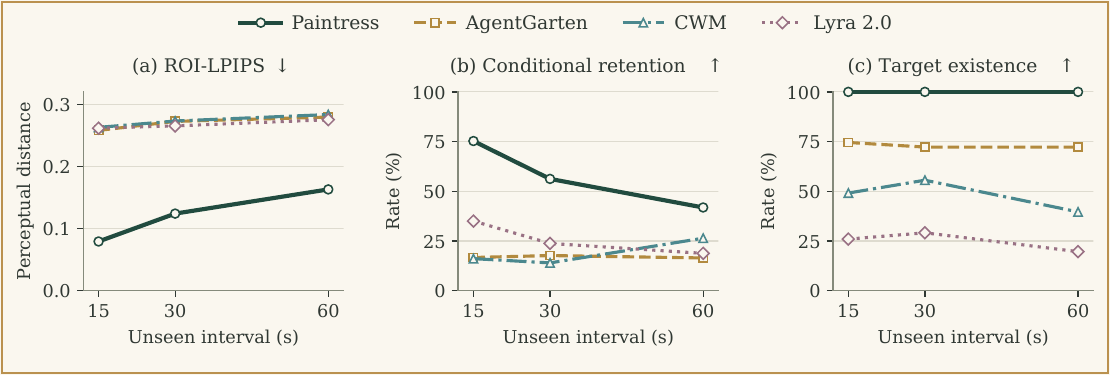}
\caption{\textbf{RQ2: Appearance discrepancy, conditional detail retention, and target existence across 15, 30, and 60\,s absence intervals ($n=55$ per point).} \method maintains 100\% target existence across all intervals and achieves lower perceptual discrepancy at 60\,s (0.1630) than the three baselines shown achieve at 15\,s ($\ge 0.2579$).}
\label{fig:degradation}
\end{figure*}

\section{Experiments and Analysis}
\label{sec:experiments}
We evaluate ten systems across all 165 settings of \bench (1,650 videos): \method, CWM~\cite{cwm}, Streaming CWM~\cite{streamingcwm}, AgentGarten~\cite{agentgarten}, GEN3C~\cite{gen3c}, SPMem~\cite{spmem}, Alaya-EVOKE~\cite{evoke}, HY-WorldPlay~\cite{hyworldplay}, Lyra~2.0~\cite{lyra2}, and Matrix-Game-4~\cite{matrixgame4}. All systems receive the same opening image, scene prompt, and camera trajectory through their native interfaces. Since backbones and training data differ, these results compare complete systems rather than isolating the memory. Five research questions examine aggregate retention (RQ1), absence horizons (RQ2), visibility timing (RQ3), semantic quality vs.\ similarity (RQ4), and runtime cost (RQ5).

\subsection{RQ1: Aggregate Appearance and Detail Retention}
\label{sec:mainresults}

Table~\ref{tab:e33_main_v06} summarizes the ten-system comparison. \method ranks first on every metric (group-balanced mean rank 1.08; Fig. 1E). The largest gap is in detail retention: \method keeps 52.12\% of target details on return (D), whereas no baseline exceeds 11.52\%.

Compared with CWM, \method reduces ROI-LPIPS by \textbf{55.3\%} and masked MAE by \textbf{56.2\%} (+8.60\,dB PSNR), while increasing overall detail retention $D$ by $4.8\times$ (52.12\% vs.\ 10.91\%) and memory gain by $3.5\times$ (0.1068 vs.\ 0.0305). Against AgentGarten and Lyra~2.0, \method lowers ROI-LPIPS by \textbf{54.8\%} and \textbf{54.4\%} (+8.39 and +10.73\,dB PSNR) and more than doubles conditional retention $R_{\mathrm{cond}}$ (57.58\% vs.\ 16.97\% and 26.06\%).

\subsection{RQ2: What Changes with Longer Absence?}
\label{sec:rq_duration}
Figure~\ref{fig:degradation} tracks the 15, 30, and 60\,s horizons ($n=55$ settings per point) for the four systems with non-trivial detail retention and timing estimability in Tables~\ref{tab:e33_main_v06}--\ref{tab:e33_timing_v05} ($R_{\mathrm{cond}} > 15\%$ and $\text{Estimable} > 25\%$: \method, AgentGarten, CWM, and Lyra~2.0; the other six baselines have $D \le 3.03\%$ or undefined $R_{\mathrm{cond}}$):
\begin{enumerate}
\item \textbf{60\,s retention surpasses the three baselines at 15\,s:} \method's ROI-LPIPS after a \textbf{60\,s} absence (\textbf{0.1630}) remains substantially lower than that all three baselines after just \textbf{15\,s} (AgentGarten 0.2579, Lyra~2.0 0.2615, CWM 0.2629), and is \textbf{40.8\%} lower than the closest baseline at 60\,s (Lyra~2.0, 0.2752). Likewise, \method's 60\,s $R_{\mathrm{cond}}$ (\textbf{41.83\%}) exceeds all three at 15\,s ($\le 35.00\%$).
\item \textbf{High-fidelity starting point vs.\ high-discrepancy regime:} Competing baselines—whether using bounded recent context (CWM, AgentGarten) or long-range geometric retrieval (Lyra~2.0)—already exhibit high discrepancy at 15\,s (0.2579--0.2629) with smaller variations between 15 and 60\,s (Lyra~2.0's $R_{\mathrm{cond}}$ declines from 35.00\% to 18.75\%; CWM's varies between 13.95\% and 26.38\%). \method starts at 0.0791 ROI-LPIPS and 75.26\% $R_{\mathrm{cond}}$ at 15\,s and drifts to 0.1630 and 41.83\% at 60\,s as longer routes accumulate intermediate observations and reconstruction steps.
\item \textbf{Persistent structure vs.\ drifting fine attributes:} \method maintains \textbf{100.00\%} target existence across all intervals (Fig.~\ref{fig:degradation}c), consistent with dual-world conditioning anchors object presence while fine surface attributes attenuate.
\end{enumerate}

\subsection{RQ3: Target-Visibility Timing}
\label{sec:rq_timing}
Table~\ref{tab:e33_timing_v05} evaluates schedule adherence. \method achieves \textbf{100.00\%} duration estimability across all 165 settings, with the lowest duration error (\textbf{[1.69, 5.75]\,s}; \textbf{[5.45\%, 18.18\%]} relative) and boundary error (\textbf{[1.04, 3.09]\,s}), and a signed bias $\Delta T$ (\textbf{[$-$1.93, $+$2.34]\,s}) centered near zero. By contrast, CWM (44.85\% estimable) skews early ($\Delta T\in[-6.17, -2.44]\,\text{s}$), AgentGarten (68.48\%) skews late ($[-0.07, +4.41]\,\text{s}$), and Lyra~2.0 (26.06\%) suffers from premature reappearance ($[-30.73, -0.64]\,\text{s}$). Matrix-Game-4 is estimable on only 5.45\% of episodes, and GEN3C, SPMem, and HY-WorldPlay yield 0.00\%.

\begin{table}[t]
\centering
\caption{\textbf{RQ4: Return visual quality across five dimensions (\%).} \emph{Exist}: target existence; \emph{Read.}: attribute readability; \emph{Sharp.}: sharpness; \emph{Clean}: no obvious artifacts; \emph{Struct.}: structural plausibility.}
\label{tab:return_quality_main}
\footnotesize
\setlength{\tabcolsep}{0pt}
\begin{tabular*}{\columnwidth}{@{\extracolsep{\fill}}lrrrrr@{}}
\toprule
System & Exist$\uparrow$ & Read.$\uparrow$ & Sharp.$\uparrow$ & Clean$\uparrow$ & Struct.$\uparrow$ \\
\midrule
\textbf{Paintress (Ours)} & \textbf{100.00} & \textbf{79.39} & \underline{81.82} & \underline{78.79} & \textbf{100.00} \\
CWM~\cite{cwm} & 47.88 & 23.64 & 30.91 & 41.21 & 28.48 \\
Streaming CWM~\cite{streamingcwm} & 32.73 & 25.45 & 47.27 & 18.79 & 13.33 \\
AgentGarten~\cite{agentgarten} & \underline{72.73} & \underline{52.73} & 67.88 & 59.39 & \underline{57.58} \\
GEN3C~\cite{gen3c} & 1.21 & 0.61 & 0.00 & 0.00 & 0.00 \\
SPMem~\cite{spmem} & 0.00 & 0.00 & 0.00 & 0.00 & 0.00 \\
Alaya-EVOKE~\cite{evoke} & 1.21 & 1.82 & \textbf{82.42} & \textbf{80.00} & 13.33 \\
HY-WorldPlay~\cite{hyworldplay} & 0.00 & 0.00 & 0.00 & 0.00 & 0.00 \\
Lyra 2.0~\cite{lyra2} & 24.85 & 24.24 & 61.21 & 51.52 & 26.06 \\
Matrix-Game-4~\cite{matrixgame4} & 6.67 & 7.88 & 45.45 & 24.24 & 5.45 \\
\bottomrule
\end{tabular*}
\end{table}

\subsection{RQ4: What Do Similarity Scores Miss?}
\label{sec:failures}
Comparing Table~\ref{tab:e33_main_v06} with Table~\ref{tab:return_quality_main} exposes three distinct failure modes:
\begin{itemize}
\item \textbf{Degenerate similarity without target synthesis:} HY-WorldPlay ranks second in ROI-LPIPS (0.1810) and PSNR (18.55\,dB), yet scores \textbf{0.00\%} across all quality dimensions (Table~\ref{tab:return_quality_main}) and timing estimability: stalled motion or uniform outputs yield small self-referenced distances despite complete task failure.
\item \textbf{Sharp rendering of the wrong place:} Alaya-EVOKE achieves the highest return sharpness (\textbf{82.42\%}) and cleanliness (\textbf{80.00\%}), yet its target existence is \textbf{1.21\%} and $D$ is \textbf{0.24\%} ($R_{\mathrm{cond}}=0.67\%$), showing that high frame quality does not imply returning to the target.
\item \textbf{Fine-detail drift under intact structure:} \method achieves \textbf{100.00\%} target existence and structural plausibility with \textbf{79.39\%} readable attributes, though $R_{\mathrm{cond}}$ declines from 75.26\% (15\,s) to 41.83\% (60\,s) due to subtle color and texture shifts.
\end{itemize}

\subsection{RQ5: What Does a Continuation Window Cost?}
\label{sec:rq_cost}
Across 21 long-history continuation windows in the 60\,s tier (37.04--68.79\,s of prior history; 90 new frames / 3.75\,s per cycle), a full cycle on $4\times\text{A800}$ (4 windows) averages \textbf{751.61\,s} (12.53\,min): 265.37\,s for memory construction (including 44.48\,s reader refinement), 159.33\,s condition encoding, 25.77\,s coverage masking, 193.79\,s four-step H3 sampling, 97.81\,s VAE decoding, and 9.53\,s publication. On $4\times\text{H20}$ (17 windows), the cycle averages \textbf{912.11\,s} (15.20\,min), with 208.10\,s for memory construction (28.47\,s reader refinement) and 351.44\,s for H3 sampling. Thus, four-step diffusion sampling accounts for only \textbf{25.8\%} (A800) to \textbf{38.5\%} (H20) of total latency, while 3D reconstruction, Gaussian fitting, and multi-process encoding/decoding dominate offline runtime.

\section{Conclusion}
We introduced \method, a code world model that couples an executable geometric proxy with an independently parameterized 3D Gaussian appearance memory via confidence-gated raw fusion and a separately refined reader copy, alongside \bench, a 165-setting benchmark evaluating matched-view return consistency against each model's own generated first visit. Across ten systems, \method improves target-region appearance and detail retention while the benchmark exposes residual attribute drift over longer absences and the runtime cost of the offline pipeline, making generated-appearance retention a concrete, measurable objective for code-conditioned world models.

{\small
\setlength{\baselineskip}{10.5pt}

}
\end{document}